\documentclass[runningheads]{llncs}

\usepackage[a4paper, margin=3cm]{geometry}%

\usepackage{graphicx}
\usepackage{layouts}
\usepackage{nicefrac}
\usepackage{url}
\usepackage{tikz}
\usetikzlibrary{decorations.pathreplacing, positioning, shapes, arrows.meta, intersections}
\tikzset{>={Latex[length=1.5mm, width=1mm]}}
\usepackage{fontawesome5}
\usepackage{booktabs}
\usepackage{subcaption}

\usepackage{xcolor}
\definecolor{d61green}{RGB}{48,184,136}
\definecolor{d61blue}{RGB}{0,169,206}
\definecolor{d61plum}{RGB}{109, 32, 119}
\definecolor{d61oceanblue}{RGB}{0, 75, 135}
\definecolor{myred}{RGB}{200, 40, 40}

\usepackage{hyperref}
\hypersetup{pdfborder={0 0 0},colorlinks=true,urlcolor=blue,linkcolor=blue,citecolor=blue,bookmarks=true}

\newif\ifblindReviewRedaction
\blindReviewRedactionfalse

\makeatletter
\newcommand\thefontsize[1]{{#1 The specified font size is: \f@size pt}}
\makeatother

\begin{document}
    %
%    \title{Deep Learning for Prawn Farming: Forecasting and Anomaly Detection}
    \title{Deep Learning for Prawn Farming:}
    \subtitle{Forecasting and Anomaly Detection}
    %
    %\titlerunning{Abbreviated paper title}
    % If the paper title is too long for the running head, you can set
    % an abbreviated paper title here
    %
    \ifblindReviewRedaction
    \author{
        Anonymous\inst{1} 
    }
    \authorrunning{Anonymous author}
    % First names are abbreviated in the running head.
    % If there are more than two authors, 'et al.' is used.
    %
    \else
        \author{
            Joel Janek Dabrowski\thanks{Corresponding author}\inst{1}
            \and
            Ashfaqur Rahman\inst{1} 
            \and
            Andrew Hellicar\inst{1}
            \and
            Mashud Rana\inst{1}$^{^*}$
            \and
            Stuart Arnold\inst{3}
        }
        \authorrunning{J.J. Dabrowski et al.}
        % First names are abbreviated in the running head.
        % If there are more than two authors, 'et al.' is used.
        %
        \institute{
            Data61, CSIRO, Australia \\
            \email{Firstname.Lastname@data61.csiro.au} \\
            $^*$ \email{Mdmashud.Rana@data61.csiro.au}
%            \and
%            Data61, CSIRO, Australia \\
%            \email{Mdmashud.Rana@data61.csiro.au}
            \and
            Agriculture \& Food, CSIRO, Australia \\
            \email{Stuart.Arnold@csiro.au}
        }
    \fi

    \maketitle              % typeset the header of the contribution
    \begin{abstract}
        We present a decision support system for managing water quality in prawn ponds. The system uses various sources of data and deep learning models in a novel way to provide 24-hour forecasting and anomaly detection of water quality parameters. It provides prawn farmers with tools to \textit{proactively} avoid a poor growing environment, thereby optimising growth and reducing the risk of losing stock. This is a major shift for farmers who are forced to manage ponds by \textit{reactively} correcting poor water quality conditions. To our knowledge, we are the first to apply Transformer as an anomaly detection model, and the first to apply anomaly detection in general to this aquaculture problem. Our technical contributions include adapting ForecastNet for multivariate data and adapting Transformer and the Attention model to incorporate weather forecast data into their decoders. We attain an average mean absolute percentage error of 12\% for dissolved oxygen forecasts and we demonstrate two anomaly detection case studies. The system is successfully running in its second year of deployment on a commercial prawn farm.
                
    \end{abstract}
    %
 
    % Author: Joel Janek Dabrowski
% Manuscript: Anomaly Detection for Prawn Farming
% Section: Introduction

\section{Introduction}

The global trade of prawn (shrimp) is estimated at 28 billion US dollars per annum and this market continues to grow at a rate faster than any other aquaculture species \cite{fao2020towards}. The main challenge in prawn farming is to manage the highly variable water quality in prawn ponds to optimise prawn health and growth \cite{boyd1998pond}.

Dissolved oxygen (DO) is generally accepted as the most important water quality parameter in aquaculture \cite{robertson2006Australian}. Excessively low values in the diurnal cycle of DO (commonly referred to as a ``DO crash'') can cause the prawn to experience hypoxia, anoxia, or death. An entire crop (typically 8 to 12 tons of prawn) can be lost in a matter of hours \cite{robertson2006Australian}. 

Farmers typically monitor water quality parameters using sensors which, especially under continuous monitoring conditions, are be subject to high levels of biofouling and harsh conditions. Biofouling can reduce a sensor's accuracy or damage it (these can cost over US\$25,000). Maintaining water quality sensors can thus be a challenging task.

This work is impactful as it provides forecasting and anomaly detection tools to assist a farmer in taking a \textit{proactive} pond management approach rather than a \textit{reactive} approach of correcting poor conditions. Forecasting gives an indication on how the temporal dynamics of a variable are expected to evolve into the future. Anomaly detection provides a means to identify any changes in the dynamics of a variable that are unusual, such as DO crashes and biofouling. With better control over water quality, animal stress can be reduced to improve growth, survivability, and consequently, production \cite{boyd1998pond}.

The novelty of this study is in the way that we provide a combination of forecasting and anomaly detection for decision support for this particular domain. Our novel technical contributions include (1) applying Transformer \cite{vaswani2017attention} for anomaly detection for the first time in the literature, (2) extending ForecastNet \cite{Dabrowski2020ForecastNet} into a multivariate model, and (3) proposing a novel approach to incorporate weather forecast data into the decoders of the Transformer \cite{vaswani2017attention} and Attention \cite{bahdanau2014neural} models in a forecasting context. This work additionally provides insight into the aquaculture domain and its challenges.

Although this work is demonstrated with prawn farming, it is applicable to other domains such as other aquaculture farming industries (such as fish, molluscs, and other crustaceans), reservoir monitoring, lake monitoring, river monitoring, coastal water monitoring, and sewer monitoring. 

    % Author: Joel Janek Dabrowski
% Manuscript: Anomaly Detection for Prawn Farming
% Section: Related Work

\section{Related Work}
\label{sec:relatedWork}

There are many challenges in precision agriculture, which have attracted various decision support tools, of which, water quality decision tools are the most common. These tools may provide decision support for determining optimal ranges of water quality \cite{Tobing2019Design} or provide sensing infrastructure \cite{Encinas2017Design}. Various physical or chemical models have also been developed, often for scenario analysis \cite{Ernst2000AquaFarm}. Various water quality forecasting approaches have been also developed \cite{Dabrowski2018State}. Anomaly detection has been applied in water quality applications (e.g., \cite{eustace2019survey}). However, to our knowledge, it has not been applied to aquaculture. 

Our system makes use of deep learning models for forecasting and anomaly detection. Temporal deep learning models are usually based on recurrent neural networks (RNNs) and convolutional neural networks (CNNs). These include the sequence-to-sequence (seq2seq) model \cite{Sutskever2014Sequence}, the attention model \cite{bahdanau2014neural}, and DeepAnT \cite{Munir2019DeepAnT}. However, there are also models that are not based on RNNs or CNNs, such as the Transformer model \cite{vaswani2017attention} and ForecastNet \cite{Dabrowski2020ForecastNet}. To our knowledge, Transformer has not been applied for anomaly detection before. According to a recent review on anomaly detection \cite{Pang2021Deep}, we consider ``generic normality feature learning'' anomaly detection approaches. 

    % Author: Joel Janek Dabrowski
% Manuscript: Anomaly Detection for Prawn Farming
% Section: Introduction

\section{System Architecture  and Overview}
\label{sec:systemArchitecture}

The decision support system architecture comprises 5 YSI EXO2 Multiparameter Sonde water quality sensors\footnote{\url{https://www.ysi.com/exo2}}, an ATMOS41 weather station\footnote{\url{https://www.metergroup.com/environment/products/atmos-41-weather-station/}}, the Senaps platform \cite{mac2017senaps}, and a server containing a website interface and the models.

Data are collected by the water quality and weather station sensors and are uploaded to the Senaps platform \cite{mac2017senaps} via the local mobile (cellular) network. Senaps organises and stores this data. Weather forecasts are obtained from the Australian Bureau of Meteorology (BoM).

The server is a virtual box comprising two 2.60GHz Intel\textsuperscript{\textregistered} Xeon\textsuperscript{\textregistered} CPUs and 8GB memory. Every hour, the server collects the latest sensor data from Senaps and the latest air temperature forecasts from BoM. The data are pre-processed for each of the machine learning models and the models are run. This entire process takes approximately 6 minutes, which provides sufficient time to provide the hourly updates and the scope to increase the update frequency.

\begin{figure}[!tb]
    \centering
    \begin{subfigure}[t]{2.8in}
        \ifblindReviewRedaction
        \includegraphics[width=2.8in]{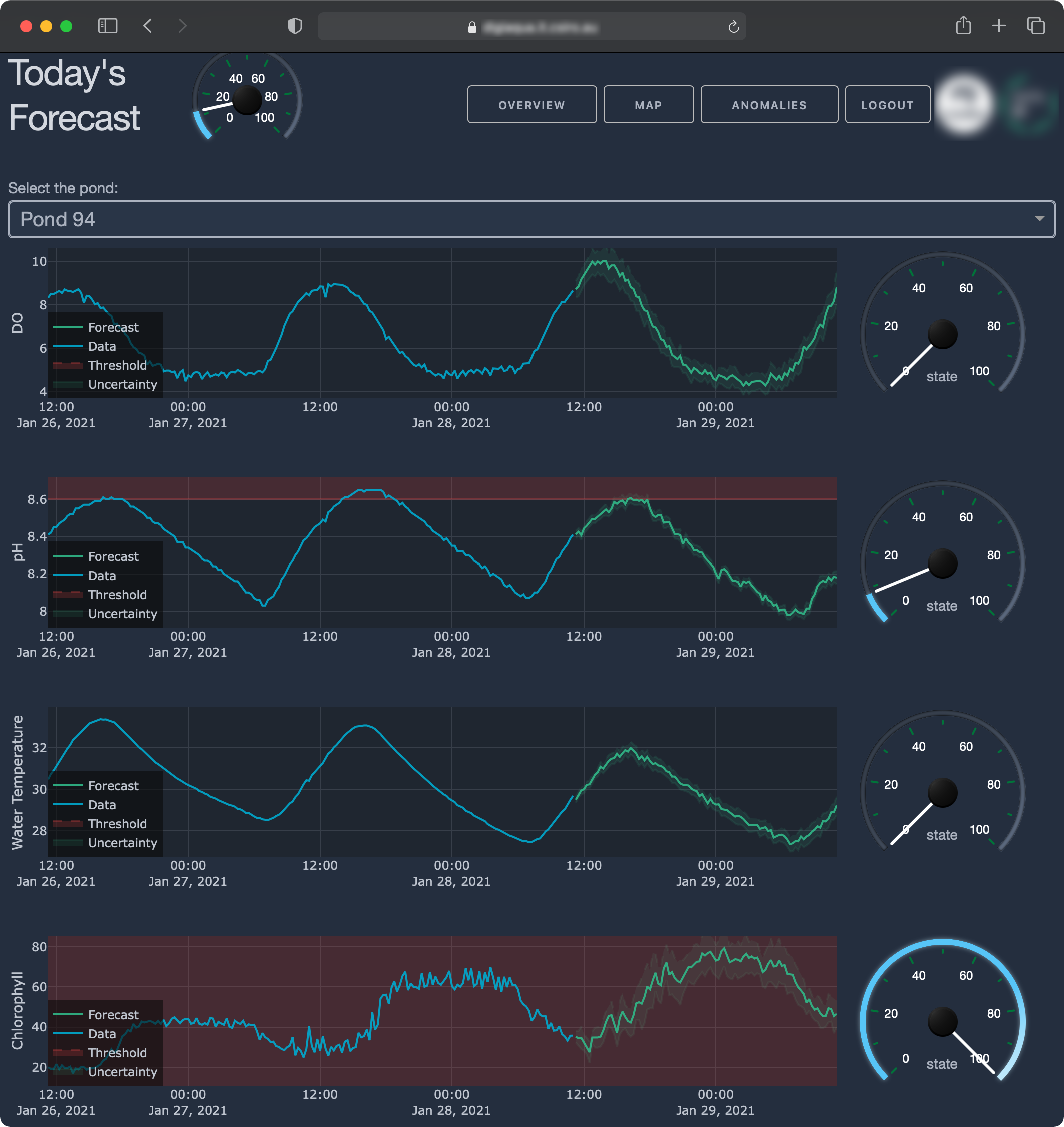}
        \else
        \includegraphics[width=2.8in]{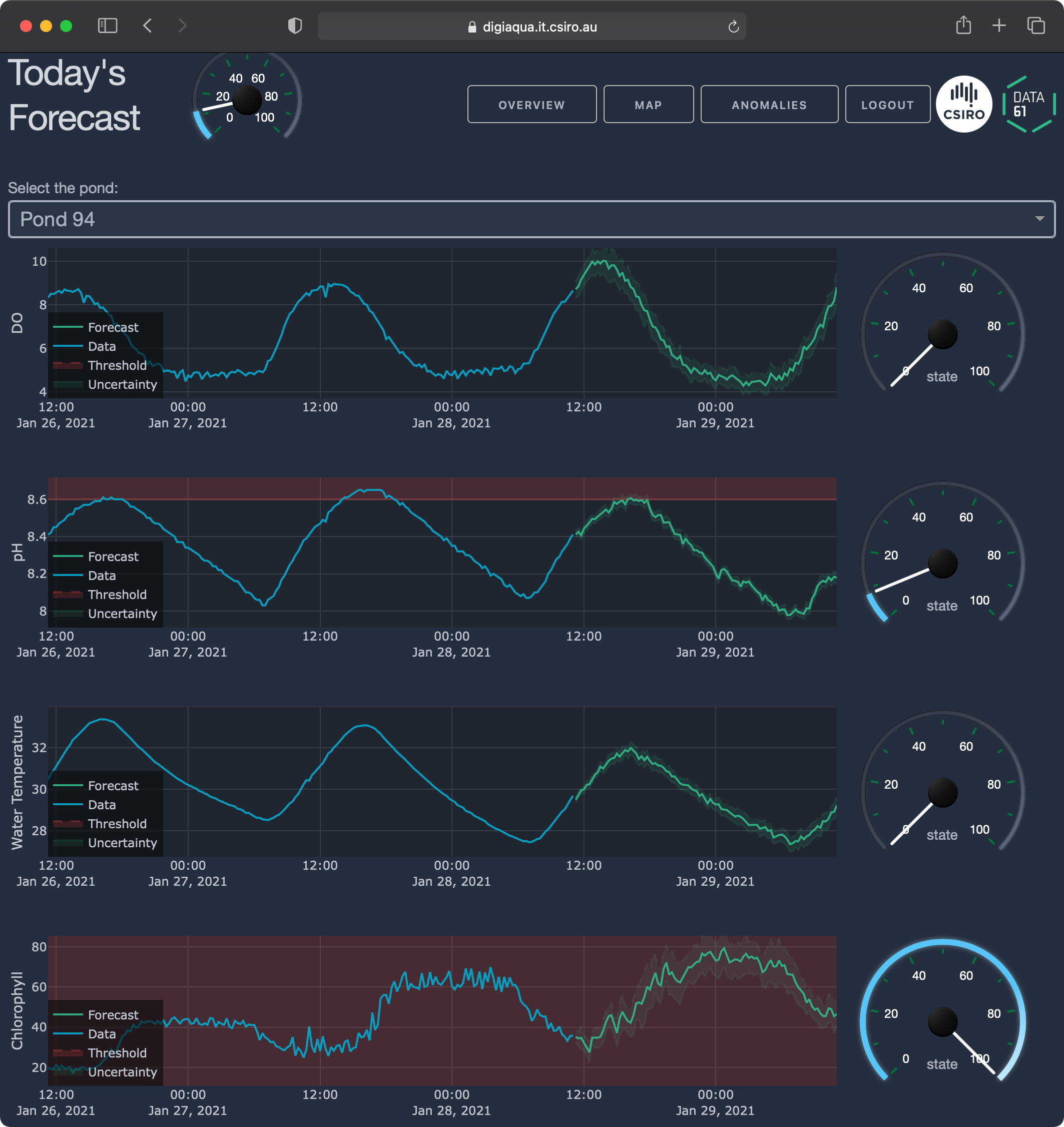}
        \fi
        \caption{Forecasts page}
        \label{fig:forecast_page}
    \end{subfigure} ~
    \begin{subfigure}[t]{2.8in}
        \ifblindReviewRedaction
        \includegraphics[width=2.8in]{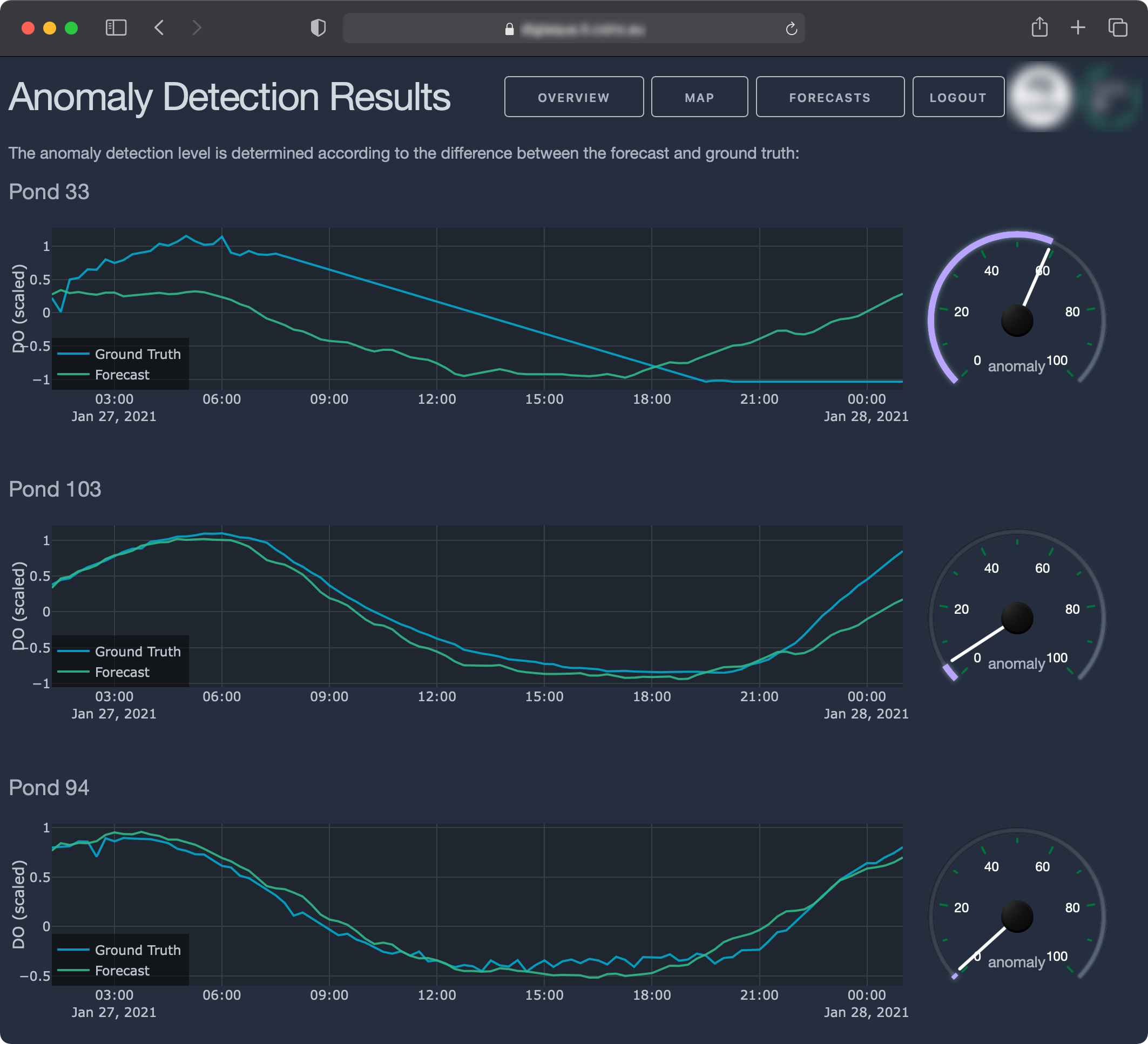}
        \else
        \includegraphics[width=2.8in]{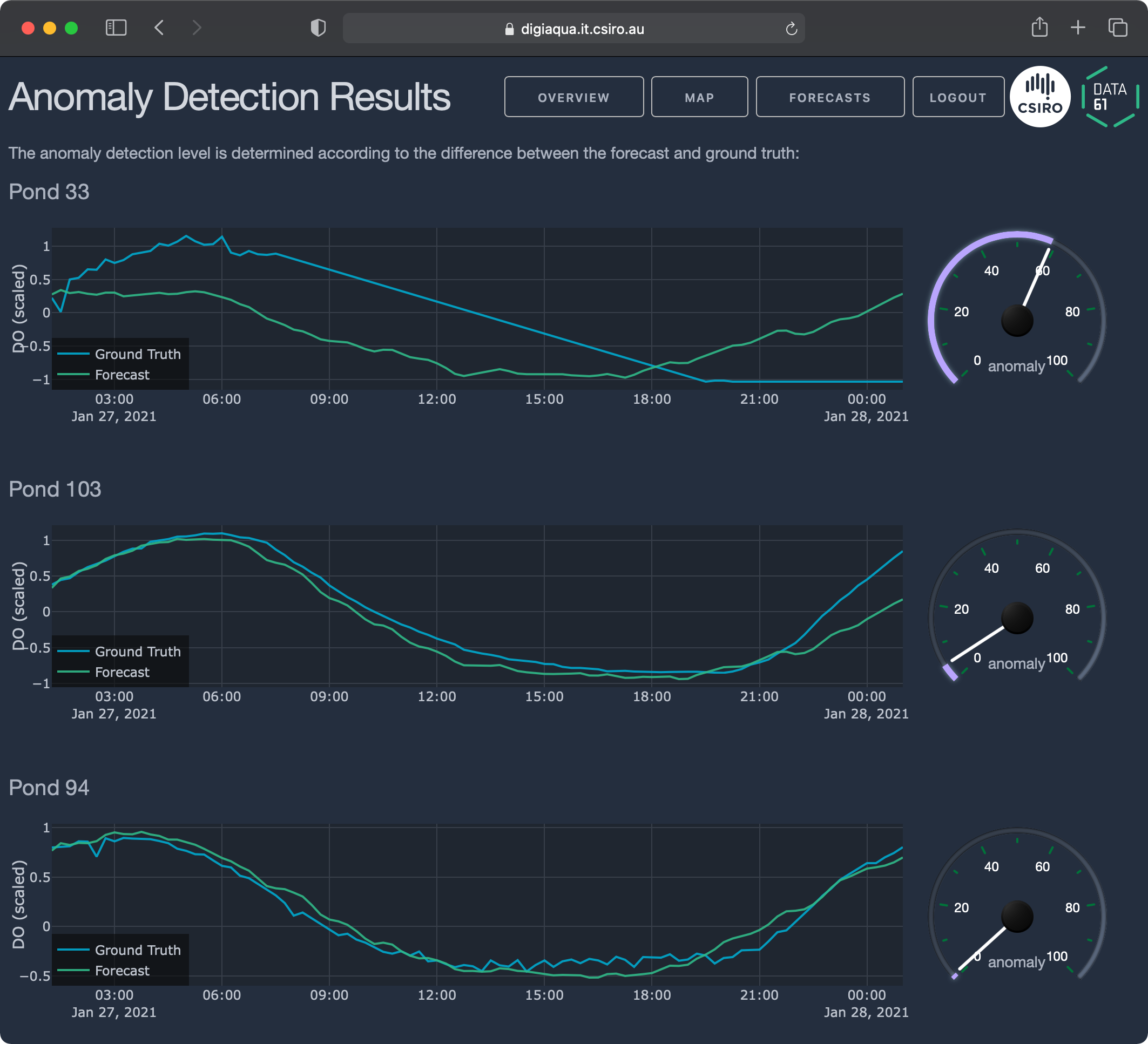}
        \fi
        \caption{Anomalies page}
        \label{fig:anomalies_page}
    \end{subfigure}
    \caption{Website application pages. The blue plots indicate the data and the green plots indicate the forecast/prediction. The grey filled region indicates forecast uncertainty and the red regions indicate threshold values. The gauges indicate the state of the water quality parameter (a) or the anomaly level (b).}
\end{figure}

\label{sec:websiteApplication}
The website application is developed on the Plotly Dash platform
\footnote{\url{https://plot.ly}}
as it provides a seamless integration with the Python code used for model development and data processing. It also provides a website interface which is accessible from computers, tablets, and smart phones.

The website contains several pages which provide navigation, plots of forecasts, and anomaly detection results. The page displaying the forecasts for DO, pH, water temperature, and chlorophyll is illustrated in \figurename{~\ref{fig:forecast_page}}. A pond ``state'' is indicated by a gauge to the right of the forecast plots. This guage provides an indication of how much and how long a forecast is expected to exceed a threshold. The gauge values are calculated according to the area of the region between the forecast curve and the horizontal threshold line, where the forecast exceeds the threshold. This area is normalised for a threshold of 180 minutes (3 hours) and maximum signal values obtained from the dataset. 

The gauge at the top left in \figurename{~\ref{fig:forecast_page}} provides an \textit{overall} estimation of the pond state. It is calculated as a weighted sum of the values for the other gauges where weighting is determined by expert knowledge from prawn farmers.

The page displaying the anomalies is illustrated in \figurename{~\ref{fig:anomalies_page}}. The plots provide the data and the anomaly detection model predictions of the data. The anomaly gauge on the right-hand side of the plot provides an anomaly score by thresholding the Mean Squared Error (MSE) between the data and prediction to a value of $1.026$. This threshold corresponds to the $99\%$ percentile of the training error produced by the anomaly detection model. That is, an MSE $\geq 1.026$ saturates to an anomaly level of $100$ on the gauge, which corresponds to a MSE that falls above the $99\%$ of MSEs achieved in the training set. 

    % Author: Joel Janek Dabrowski
% Manuscript: Anomaly Detection for Prawn Farming
% Section: Methods

\section{Methods}
\label{sec:methods}

\subsection{Forecasting}

With a diurnal cycle in DO, the forecasting models are designed to use 48-hours of historical data to provide a 24-hour, multi-step-ahead forecast. The inputs and outputs of the models are selected according to expert knowledge. The inputs include DO, pH, chlorophyll, water temperature, air temperature, and an air temperature forecast. The outputs of the models provide a forecast of DO, pH, water temperature, or chlorophyll. A different model is constructed and trained for each variable that is forecast. The forecasting models considered comprise, Transformer \cite{vaswani2017attention}, Attention \cite{bahdanau2014neural}, and ForecastNet \cite{Dabrowski2020ForecastNet}.

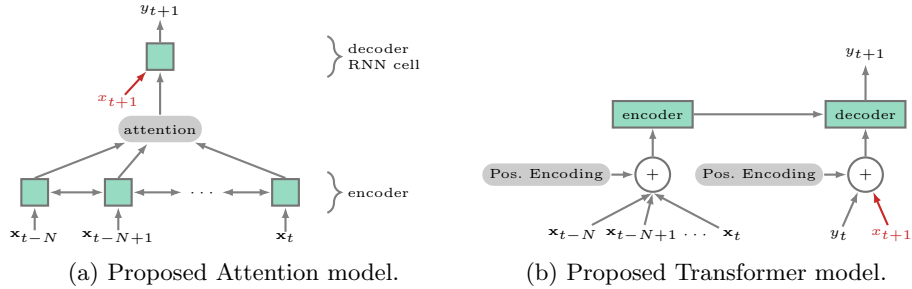
\begin{figure}[!tb]
    \centering
    \begin{subfigure}[t]{2.32in}
        %Joel Dabrowski
%Tikz Figure
%Name: fig_attention1
%Basic overview of the attention mechanism

\def\horisep{1.1cm}
\def\vertsep{0.6cm}

\begin{tikzpicture}[->,draw=black!50, thick]
    \begin{tiny}
        \tikzstyle{cell}=[rectangle, draw=black!55, fill=d61green!50, minimum size=10pt, inner sep=0pt]
        \tikzstyle{operator}=[rounded rectangle,fill=black!20, minimum width=\horisep, minimum height=10pt,inner sep=0pt]
        \tikzstyle{input}=[rectangle, inner sep=0pt]
        
%        \node[input] (o1) at (1*\horisep,4*\vertsep) {$y_{t+1}$};
        \node[input] (o2) at (2.5*\horisep,4*\vertsep) {$y_{t+1}$};
%        \node[input] (o3) at (4*\horisep,4*\vertsep) {$y_{t+3}$};
        
        %Decoder RNN
%        \node[cell] (s1) at (1*\horisep,3*\vertsep) {};
        \node[cell] (s2) at (2.5*\horisep,3*\vertsep) {};
%        \node[cell] (s3) at (4*\horisep,3*\vertsep) {};
%        \path (s1) edge (s2);
%        \path (s2) edge (s3);
        
%        \path (s1) edge (o1);
        \path (s2) edge (o2);
%        \path (s3) edge (o3);
        
%        \node[input] (i1p) at (1*\horisep,2*\vertsep) {$x_{t+1}$};
        \node[input] (i2p) at (2*\horisep,2*\vertsep) {\textcolor{myred}{$x_{t+1}$}};
%        \node[input] (i3p) at (4*\horisep,2*\vertsep) {$x_{t+3}$};
%        \path[] (i1p) edge (s1);
        \path[draw=myred] (i2p) edge (s2);
%        \path[] (i3p) edge (s3);
        
        %Encoder RNN
%        \node[cell, label={[yshift=-8pt]right:$y_{t-N}$}] (h1) at (1*\horisep,0*\vertsep) {};
%        \node[cell, label={[yshift=-8pt]right:$y_{t-N+1}$}] (h2) at (2*\horisep,0*\vertsep) {};
%        \node[]     (ht) at (3*\horisep,0*\vertsep) {$\cdots$};
%        \node[cell, label={[yshift=-8pt]right:$y_{t}$}] (hT) at (4*\horisep,0*\vertsep) {};
        \node[cell] (h1) at (1*\horisep,0*\vertsep) {};
        \node[cell] (h2) at (2*\horisep,0*\vertsep) {};
        \node[]     (ht) at (3*\horisep,0*\vertsep) {$\cdots$};
        \node[cell] (hT) at (4*\horisep,0*\vertsep) {};
        \path[<->] (h1) edge (h2);
        \path[<->] (h2) edge (ht);
        \path[<->] (ht) edge (hT);
        
        \node[input] (i1) at (1*\horisep,-1*\vertsep) {$\mathbf{x}_{t-N}$};
        \node[input] (i2) at (2*\horisep,-1*\vertsep) {$\mathbf{x}_{t-N+1}$};
        \node[input] (i3) at (4*\horisep,-1*\vertsep) {$\mathbf{x}_{t}$};
        \path[] (i1) edge (h1);
        \path[] (i2) edge (h2);
        \path[] (i3) edge (hT);

        %Draw sum of alpha_{i,j} and h_j
        \node[operator] (sum) at (2.5*\horisep,1.4*\vertsep) {attention};
        
        %Draw product of alpha_{i,j} and h_j
%        \draw (h1.north) -> node[anchor=east, xshift=-0.1cm] {$\alpha_{t,1} a_{\text{enc}}^{<1>}$} (sum);
%        \draw (h2.north) -> node[anchor=west, xshift=-0.1cm] {$\alpha_{t,2} a_{\text{enc}}^{<2>}$} (sum);
%        \draw (hT.north) -> node[anchor=west, xshift=0.1cm] {$\alpha_{t,T} a_{\text{enc}}^{<T>}$} (sum);
        \draw (h1.north) -> (sum);
        \draw (h2.north) -> (sum);
        \draw (hT.north) -> (sum);
        
%        \draw [-,decorate,decoration={brace, amplitude=5pt,raise=0pt},yshift=0pt] (5*\horisep,4.3*\vertsep) -- (5*\horisep,3.55*\vertsep) node [black,midway,xshift=5pt, right] {forecasts};
        \draw [-,decorate,decoration={brace, amplitude=5pt,raise=0pt},yshift=0pt] (4.5*\horisep,3.45*\vertsep) -- (4.5*\horisep,2.55*\vertsep) node [black,midway,xshift=5pt,right, align=left] {decoder \\ RNN cell};
%        \draw [-,decorate,decoration={brace, amplitude=5pt,raise=0pt},yshift=0pt] (5*\horisep,2.45*\vertsep) -- (5*\horisep,1.7*\vertsep) node [black,midway,xshift=5pt, right] {air temp. forecast};
%        \draw [-,decorate,decoration={brace, amplitude=5pt,raise=0pt},yshift=0pt] (5*\horisep,1.45*\vertsep) -- (5*\horisep,0.55*\vertsep) node [black,midway,xshift=5pt, right] {attention};
        \draw [-,decorate,decoration={brace, amplitude=5pt,raise=0pt},yshift=0pt] (4.5*\horisep,0.45*\vertsep) -- (4.5*\horisep,-0.45*\vertsep) node [black,midway,xshift=5pt, right] {encoder};
%        \draw [-,decorate,decoration={brace, amplitude=5pt,raise=0pt},yshift=0pt] (5*\horisep,-0.55*\vertsep) -- (5*\horisep,-1.3*\vertsep) node [black,midway,xshift=5pt, right] {sensor data};

        \draw (sum) -> (s2);
%        \path (sum) edge (s2);
    \end{tiny}
\end{tikzpicture}
        \caption{Proposed Attention model. }
        \label{fig:attention_model}
    \end{subfigure} ~
    \begin{subfigure}[t]{2.32in}
        %Joel Dabrowski
%Tikz Figure
%Name: fig_attention1
%Basic overview of the attention mechanism

\def\horisep{0.7cm}
\def\vertsep{0.8cm}

\begin{tikzpicture}[->,draw=black!50, thick]
    \begin{tiny}
        \tikzstyle{encoder}=[rectangle, draw=black!55, fill=d61green!50, minimum width=1.5*\horisep, minimum height=10pt, inner sep=3pt]
        \tikzstyle{decoder}=[rectangle, draw=black!55, fill=d61green!50, minimum width=1.5*\horisep, minimum height=10pt, inner sep=3pt]
        \tikzstyle{posenc}=[rounded rectangle, fill=black!20, inner sep=2pt]
        \tikzstyle{cell}=[rectangle, draw=black!55, fill=d61green!50, minimum width=1.5*\horisep, minimum height=20pt, inner sep=3pt]
        \tikzstyle{operator}=[circle,draw=black!55,minimum size=10pt]
        \tikzstyle{input}=[rectangle, inner sep=1pt]
        
        %Encoder
        \node[encoder] (enc) at (1.5*\horisep,2*\vertsep) {encoder};
        \coordinate (coor1) at (1.5*\horisep, 2*\vertsep - 5pt);
        
        \node[posenc] (posenc1) at (-0.5*\horisep,1*\vertsep) {Pos. Encoding};
        \node[operator] (emb1) at (1.5*\horisep,1*\vertsep) {+};
        \draw [] (emb1) to (coor1);
        \draw [] (posenc1) to (emb1);
        
        \node[input] (i1) at (0*\horisep,0*\vertsep) {$\mathbf{x}_{t-N}$};
        \node[input] (i2) at (1.3*\horisep,0*\vertsep) {$\mathbf{x}_{t-N+1}$};
        \node[input] (i3) at (2.3*\horisep,0*\vertsep) {$\cdots$};
        \node[input] (i3) at (3*\horisep,0*\vertsep) {$\mathbf{x}_{t}$};
        \draw [] (i1) to (emb1.south);
        \draw [] (i2) to (emb1.south);
        \draw [] (i3) to (emb1.south);

        \node[decoder] (dec) at (5.5*\horisep,2*\vertsep) {decoder};
        \coordinate (coor2) at (5.5*\horisep, 2*\vertsep - 5pt);
        \coordinate (coor3) at (5.5*\horisep, 2*\vertsep + 5pt);
        
        \node[posenc] (posenc2) at (3.5*\horisep,1*\vertsep) {Pos. Encoding};
        \node[operator] (emb2) at (5.5*\horisep,1*\vertsep) {+};
        \draw [] (emb2) to (coor2);
        \draw [] (posenc2) to (emb2);
        
        \node[input] (i4) at (5*\horisep,0*\vertsep) {$y_{t}$};
        \draw [] (i4) to (emb2);
        
        \node[input] (i5) at (6*\horisep,0*\vertsep) {\textcolor{myred}{$x_{t+1}$}};
        \draw [draw=myred] (i5) to (emb2);
        
        \node[input] (i5) at (5.5*\horisep,3*\vertsep) {$y_{t+1}$};
        \draw [] (coor3) to (i5);
        
        \draw [] (enc) to (dec);
%        
%        \node[input] (i4) at (5*\horisep,0*\vertsep) {$\mathbf{y}_{t}$};
%        \draw [] (i4) to (coor4);
%        
%        \node[input] (i5) at (5*\horisep,3.2*\vertsep) {$\mathbf{y}_{t+1}$};
%        \draw [] (coor5) to (i5);
        
%        \node[cell] (enc) at (1*\horisep,0*\vertsep) {decoder};

    \end{tiny}
\end{tikzpicture}
        \caption{Proposed Transformer model.}
        \label{fig:transformer_model}
    \end{subfigure}
    \caption{Model modifications to include weather forecasts in the decoder. Sensor data $\mathbf{x}_{t-N:t}$ are provided to the encoder. Air temperature forecasts \textcolor{myred}{$x_{t+1}$} are provided as inputs to the decoder. The decoder outputs the forecast $y_{t+1}$.}
    \label{fig:modelMods}
\end{figure}

The Transformer and Attention models are both encoder-decoder based models. The historical input data sequence is typically provided to the model's encoder and the model's decoder outputs the forecast. In this study, the historical data are all 48-hour sequences. The air temperature forecast data are however 24-hour sequences, which poses a challenge in combining the input data.

To address this, we modify these models to incorporate the air temperature forecast data into the decoder rather than the encoder. This is also more natural as it provides a better separation of the data associated with the past and the future. In the Attention model, the air temperature forecasts are concatenated with the standard decoder inputs, which include the attention context and the previous decoder RNN output. In the Transformer model, the air temperature forecast is concatenated with the shifted output that is passed through the decoder's positional encoding. The proposed Attention and Transformer model architectures are illustrated in \figurename{~\ref{fig:attention_model}} and \figurename{~\ref{fig:transformer_model}} respectively. 

ForecastNet does not make use of the encoder-decoder architecture, but it was proposed as a univariate model \cite{Dabrowski2020ForecastNet}. The inputs to the model are in the form of a vector which contains a univariate data sequence. We extend ForecastNet into a multivariate model by concatenating the univariate sequences from all data sources into a single vector. This concatenated vector serves as the multivariate input to ForecastNet. As such, the fundamental architecture of ForecastNet is left unchanged. In the deployed model, the output layer comprises a Gaussian output to provide uncertainty in the outputs. For compatibility, a linear output is used when comparing ForecastNet with other models. 

To demonstrate the effect of the model modifications proposed in this study, the standard form counterparts of the proposed models are also considered in the results. The standard forms of the models are configured with identical parameters (such as number of layers and hidden units), except the air temperature forecasts are not included in the models. The configurations for all the forecasting models are presented in Table \ref{table:model_config}. Parameters are selected by trial and error to optimise performance and memory requirements.

\begin{table}[!tb]{}
    \caption{Forecast (top) and anomaly detection (bottom) model configurations.}
    \label{table:model_config}
    \setlength\tabcolsep{2pt}
    \begin{center}
        \begin{scriptsize}
            \begin{tabular}{l p{0.87\columnwidth}}
                \toprule
                Model & Configuration \\
                \midrule
                proposed FN &  The proposed ForecastNet model \cite{Dabrowski2020ForecastNet} with hidden blocks configured with three fully connected layers, each containing 24 neurons. Outputs are linear outputs for testing, and a mixture model with a single Gaussian element in deployment. \\
                FN &  Identical to \textit{proposed FN}, but the temperature forecasts are not included in the inputs. \\
                proposed att & The proposed Attention model with the encoder and decoder configured with LSTMs, each with a hidden size of 48. Standard hyperbolic tangent alignment model is used in the attention mechanism. \\
                att &  Standard Attention model \cite{bahdanau2014neural} which is identical to \textit{proposed att}, but the temperature forecasts are not included as inputs to the decoder. \\
                proposed trans & The proposed Transformer model is configured with a single layer in the encoder and decoder, with a model dimension of 16, 4 heads, and 24 hidden units. \\
                trans &  Standard Transformer model \cite{vaswani2017attention} which is identical to \textit{proposed trans}, but the temperature forecasts are not included as inputs to the decoder. \\
                \midrule
                rnnAe & Long Short Term Memory (LSTM) autoencoder \cite{srivastava2015unsupervised} with a hidden size of 24 and sequence length of 96. \\
                deepAe & Fully connected deep autoencoder with input size of 96 and 3 hidden layers with sizes 56, 41, and 32 in the encoder and decoder. \\
                rnnAeFc & LSTM forecasting autoencoder \cite{srivastava2015unsupervised} with a hidden size of 24 and sequence length of 96. \\
                seq2seq & Sequence to sequence model \cite{Sutskever2014Sequence} with two LSTM layers each with a hidden size of 24 in the encoder and decoder. \\
                attention & Attention model \cite{bahdanau2014neural} with a hidden size of 24 in the encoder and decoder, with a hyperbolic tangent alignment model in the attention mechanism. \\
                deepAnt & DeepAnT model \cite{Munir2019DeepAnT} configured for 192 inputs and 96 outputs. \\
                transf. & Transformer model \cite{vaswani2017attention} comprising a single layer in the encoder and decoder, with a model dimension of 16, 4 heads, and 24 hidden units. \\
                forecastNet & ForecastNet model \cite{Dabrowski2020ForecastNet} with 192 inputs and 96 outputs. The hidden blocks contain two fully connected layers with 24 neurons in each layer. \\
                \bottomrule
            \end{tabular}
        \end{scriptsize}
    \end{center}
\end{table}

\subsection{Anomaly Detection}

We consider autoencoder and forecasting models for anomaly detection \cite{gamboa2017deep}. Autoencoder models make a prediction of their inputs through an encoder-decoder network, whereas forecasting models use the inputs to make a prediction into the future. Anomaly detection is achieved by comparing the predicted and observed data sequences according to the MSE. The predictive nature of the forecasting-based models encourages earlier detection of anomalies.

We consider several univariate anomaly detection models as described in Table \ref{table:model_config}. The RNN autoencoder (rnnAe) and a deep autoencoder (deepAe) provide autoencoder architectures. The remaining models are configured for forecasting-based anomaly detection (see Section \ref{sec:relatedWork}). Transformer is included as a new approach and is expected to perform well given its forecasting ability.

    % Author: Joel Janek Dabrowski
% Manuscript: Anomaly Detection for Prawn Farming
% Section: Dataset

\section{Dataset, Preprocessing, and Evaluation}
\label{sec:datasetAndEvaluation}

The dataset used in this study was acquired by sensors placed in several ponds on a prominent Australian prawn farm. All sensors were configured to sample at 15-minute intervals and were deployed in different ponds over two grow-out seasons covering the period of 2018-09-30 to 2020-04-19. YSI EXO2 Sensors were only in the ponds during grow-out periods (which differed between ponds). Some sensors were also removed for extended periods for servicing and repair.

The dataset used for testing the forecasting models is compiled from periods where all sensor data are present. The data are split into a training set and a test set, where both datasets contain data from all sensors that were available at the time. The training dataset contains 27 553 samples and the test set contains 1450 samples. To evaluate the forecasting models, the Mean Absolute Percentage Error (MAPE) and the Root Mean Squared Error (RMSE) metrics are used.

The anomaly detection methods are demonstrated with two case studies of known anomalous events that posed a significant risk to the farmer. The first is a DO crash event that occurred on 3 March 2020 for Sensor 1 and the second is a biofouling event that occurred over late December 2019 for Sensor 2.

During training and testing, a leave-one-out approach is adopted. The test set contains the data for the test sensor and the training dataset comprises data from all other sensors. The trained model is thus not conditioned on any form of normality or abnormality of the sensor used for testing. This also generalises the models across sensors. Additionally, the DO crash or biofouling regions in the dataset sequences are removed when these data are used in the training set.

The data are formatted such that the autoencoder anomaly detection model's inputs contain 24-hour (96 sample) data sequences and the forecasting model's (both the anomaly detection and forecasting models) inputs contain 48-hour (192 sample) data sequences and outputs produce 24-hour (96 sample) forecasts. The result is a set of training and testing input/output sequence pairs.

Data preprocessing involves standardisation and imputation. Missing values are imputed using linear interpolation if the missing value sequence length is 8 or less samples (2 hours), and removed if the missing sequence length is longer.

Both anomaly detection and forecasting models are trained using the Adam optimisation algorithm to optimise the MSE. A hyperparameter search is conducted for the learning rate over the range of $10^{i}, i=[-6,\dots,-2]$. Early stopping is used to address over-fitting.
    % Author: Joel Janek Dabrowski
% Manuscript: Anomaly Detection for Prawn Farming
% Section: Results

\section{Forecasting Results}
\label{sec:forecastingResults}

\begin{figure}[!tb]
    \centering
    \includegraphics[width=5.6in]{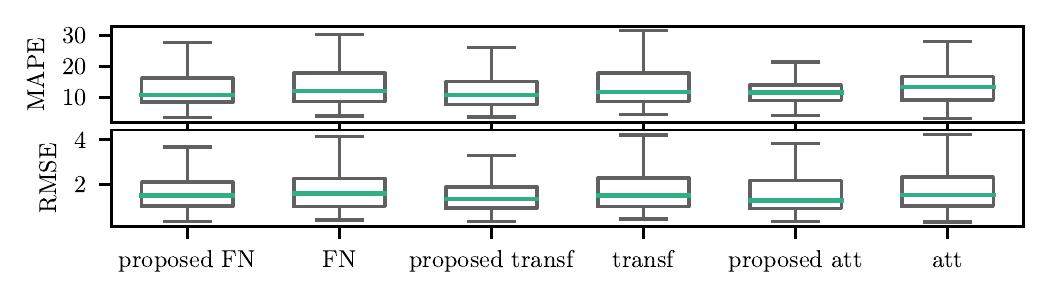}
    \caption{Test dataset forecasting error box plots for the standard and proposed ForecastNet (FN), Transformer (trans) and Attention (att) models.}
    \label{fig:box_plots}
\end{figure}

The standard and proposed ForecastNet, Transformer, and Attention models are compared with each other using the forecasting dataset. The average error results for all the models are provided in Table \ref{table:forecastResults} and box-whisker plots are provided in \figurename{~\ref{fig:box_plots}}. Additionally, the total time for a model to process the 1450 test samples is provided in the last column of Table \ref{table:forecastResults}.

Comparing the difference between the standard and the proposed versions of the models, it is clear that the forecast error and its variation are consistently reduced in the proposed models. Comparing the ForecastNet, Transformer, and Attention models, the error values do not differ significantly. Furthermore, the overlapping of box plots in \figurename{~\ref{fig:box_plots}} indicate that there is little statistical significance between these differences in error. ForecastNet is however significantly faster than the other models as indicated in Table \ref{table:forecastResults}. For example, ForecastNet is 67 times faster than the Transformer model. ForecastNet is thus deployed.

%
% Table generated by pacReefForecasting2/paperPlots.py
\begin{table}[t]{}
    \caption{Average errors over the test dataset for the standard and proposed ForecastNet (FN), Transformer (trans) and Attention (att) models. Proposed model results are presented in bold text.}
    \label{table:forecastResults}
    \begin{center}
        \begin{scriptsize}
            \setlength{\tabcolsep}{5pt}
            \begin{tabular}{l c c c c c}
                \toprule
                Model & MAPE & MAPE std dev. & RMSE & RMSE std dev. & run time (s) \\
                \midrule
                proposed FN & \textbf{12.259} & \textbf{5.181} & \textbf{1.665} & \textbf{0.821} & \textbf{0.84} \\
                FN & 13.248 & 5.665 & 1.722 & 0.842 & 0.83 \\
                \specialrule{0.2pt}{0pt}{1pt}
                proposed transf & \textbf{11.844} & \textbf{4.954} & \textbf{1.513} & \textbf{0.768} & \textbf{55.57} \\
                transf & 13.593 & 6.125 & 1.731 & 0.884 & 59.82 \\
                \specialrule{0.2pt}{0pt}{1pt}
                proposed att & \textbf{12.059} & \textbf{4.430} & \textbf{1.595} & \textbf{0.828} & \textbf{22.60} \\
                att & 13.663 & 5.786 & 1.713 & 0.821 & 22.24 \\
                \bottomrule
            \end{tabular}
        \end{scriptsize}
    \end{center}
\end{table}

\section{Anomaly Detection Case Studies}

\subsection{DO Crash Case Study}

\begin{figure}[!tb]
    \centering
    \includegraphics[width=5.6in]{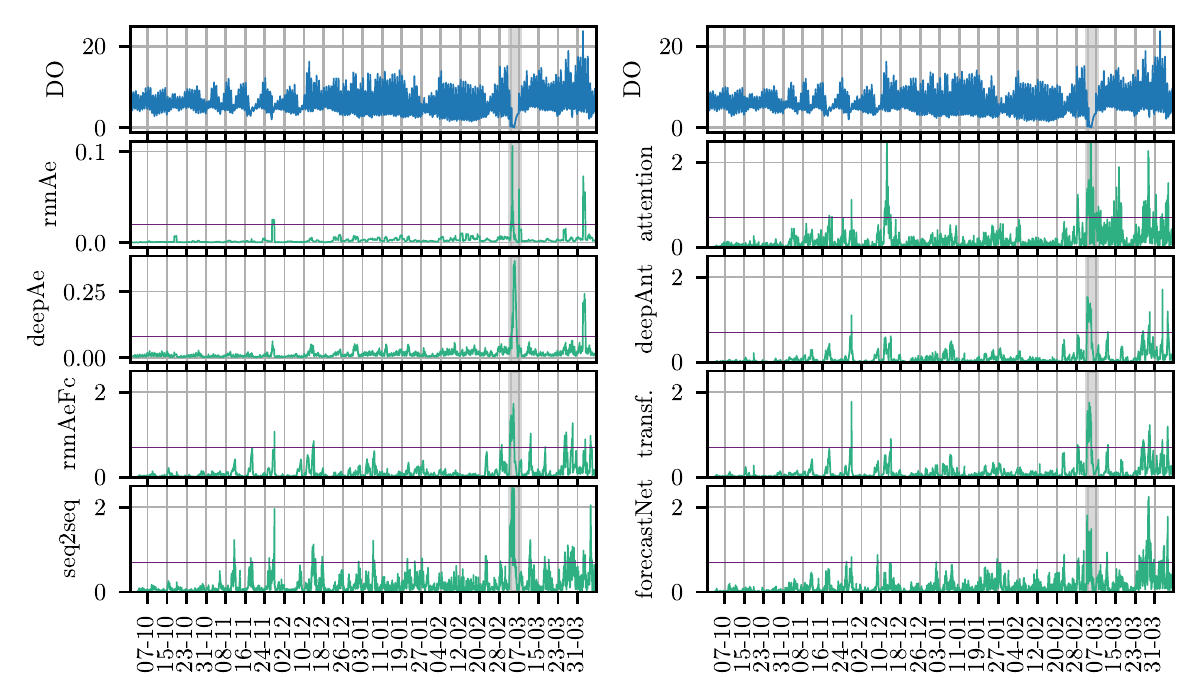}
    \caption{Anomaly detection results for Sensor 1. The DO data (mg/L) are plotted in the top panels in blue, and the MSE for each model are plotted in the remaining panels in green. The purple horizontal lines plot the $0.7$ threshold for anomaly detection. The grey region indicates the approximate period over which the DO crash occurred. Horizontal axes indicate the day and month.}
    \label{fig:anomaly_detection_1}
\end{figure}

\begin{figure}[!tb]
    \centering
    \includegraphics[width=5.6in]{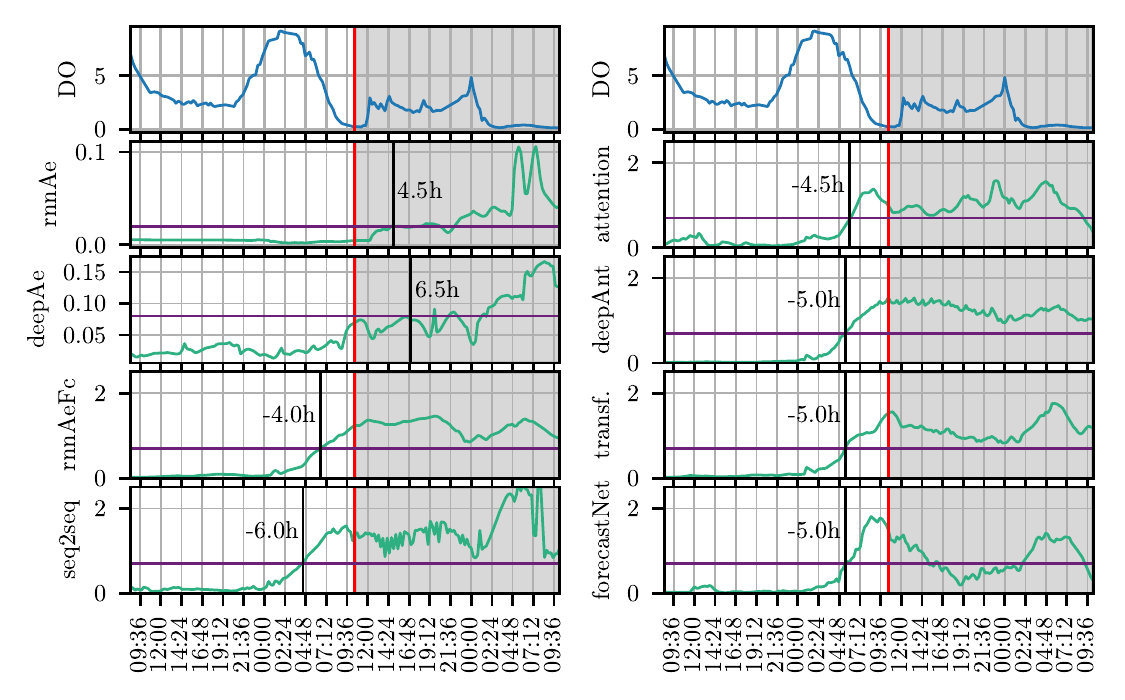}
    \caption{DO crash for sensor 1. The DO data (mg/L) are plotted in the top panels in blue, and the MSE for each model are plotted in the remaining panels in green. The purple horizontal lines plot the $0.7$ threshold for anomaly detection. The red vertical lines indicate the time at which the DO reaches its lowest point. Black vertical lines and labels indicate the time (hours) where the anomaly detection threshold is crossed. Horizontal axes indicate time.}
    \label{fig:do_detection}
\end{figure}

The anomaly detection MSE results for Sensor 1 over the 2019/2020 season are plotted in \figurename{~\ref{fig:anomaly_detection_1}}. As discussed in Section \ref{sec:websiteApplication}, the maximum anomaly detection level of 100 corresponds with a MSE of $1.026$. Suppose a farmer chooses to investigate an anomaly when the anomaly detection level rises above a value of $70$. This value corresponds to a MSE threshold of $0.7$. A plot of the DO crash, anomaly detection MSE, and threshold value are presented in \figurename{~\ref{fig:do_detection}}. Additionally, the number of hours warning that the anomaly detection provides for the DO crash is also indicated. The number of hours is measured from where the MSE exceeds the threshold to where the DO reaches its minimum. 

%
% See pacReefAnomalyDetection/paperPlots.py

All models detect the DO crash on 03-03-2020. The RNN autoencoder and deep autoencoder models detect the DO crash and have relatively few anomaly detections. However, these models detect the DO crash only after it occurs. As these are autoencoder models, they do not have the predictive nature that the forecasting anomaly detection models have.

The seq2seq model provides the earliest detection, however it has a relatively noisy MSE signal suggesting that it is more likely to make false positive detections. The Attention model provides similar noisy characteristics. The deepAnT and Transformer models provide strong detections of the DO crash with relatively low noise. ForecastNet provides early detection, however the detection weak. As ForecastNet is time-variant, it may be able to adapt better to the anomalous events resulting in less sensitivity to these anomalies.

If the farmer were to choose a lower threshold, earlier detections could be achieved at cost of more false positives. The choice of a threshold value is left to the discretion of the farmer as it will depend on their particular management practices, farm size, and farm resources.

\subsection{Sensor Biofouling Case Study}

\begin{figure}[!tb]
    \centering
    \includegraphics[width=2.0in]{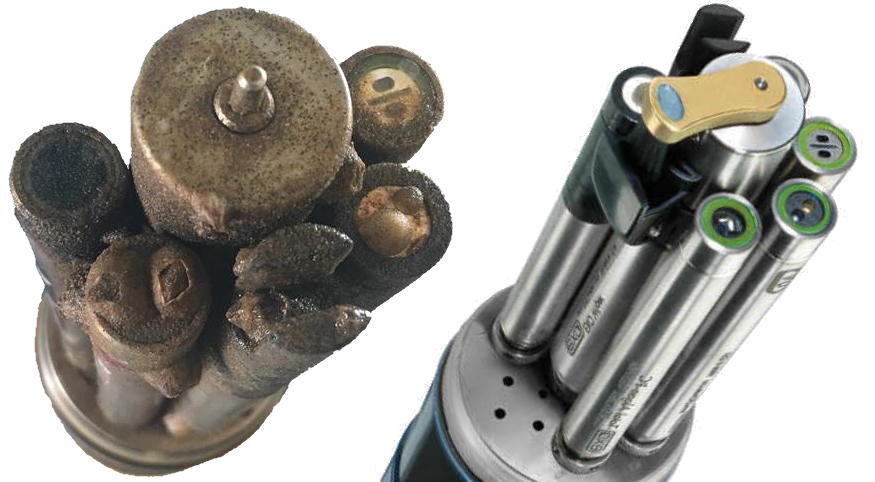}
    \caption{Sensor 2 with severe biofouling on the left and a new sensor on the right. Notice the micro-organisms and barnacles growing on the sensor heads.}
    \label{fig:biofouling}
\end{figure}

\begin{figure*}[!tb]
    \centering
    \includegraphics[width=5.6in]{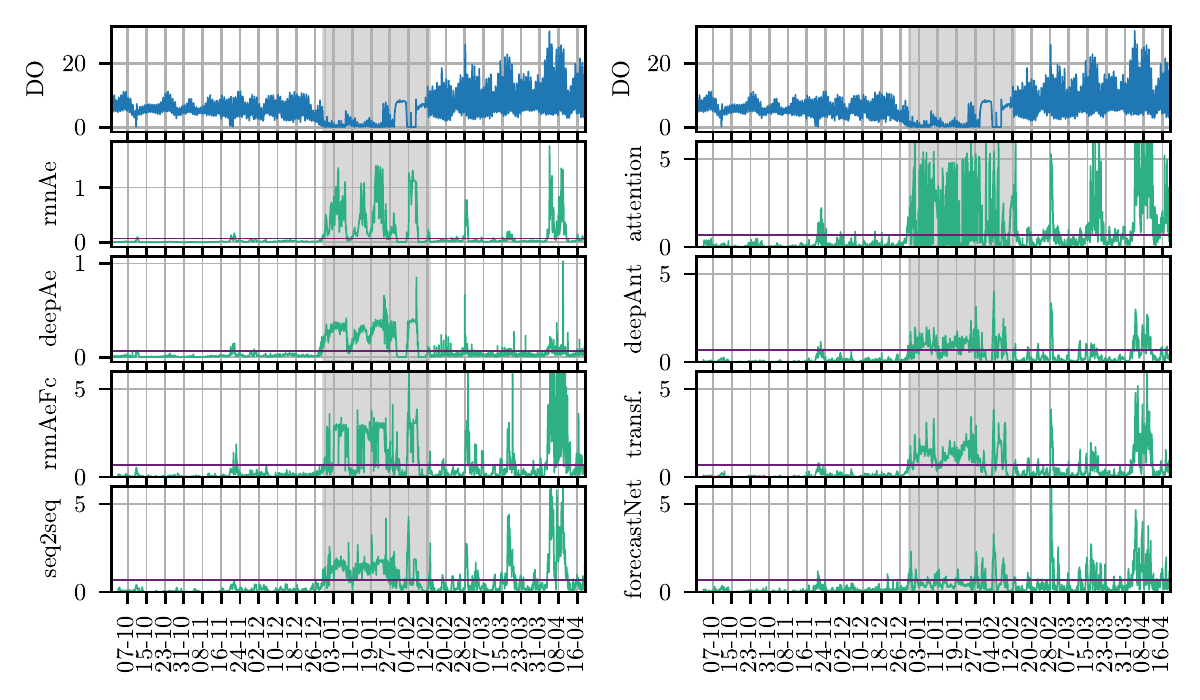}
    \caption{Anomaly detection results for Sensor 2. The DO data (mg/L) are plotted in the top panels in blue, and the MSE for each model are plotted in the remaining panels in green. The purple horizontal lines plot the $0.7$ threshold for anomaly detection. The grey region indicates the approximate period over which the sensor experienced biofouling. Horizontal axes indicate the day and month.}
    \label{fig:anomaly_detection_2}
\end{figure*}

The anomaly detection MSE results for the deployment period for the 2019/2020 season for Sensor 2 is plotted in \figurename{~\ref{fig:anomaly_detection_2}}. We were notified on 20-01-2020 by the farm that the sensor had (unintentionally) not been maintained. It is not clear when the sensor was last serviced. From anomaly detection model results, it appears the biofouling began to affect the sensor from late December 2020. The sensor was removed from the pond, cleaned, tested, and returned to the pond on 12-02-2020. A photograph of the bio-fouled sensor is provided in \figurename{~\ref{fig:biofouling}}.

As indicated in \figurename{~\ref{fig:anomaly_detection_2}}, all anomaly detection models detect the biofouling event. Furthermore, the model's anomaly detection levels remain high throughout the duration of the biofouling event. The detections made by Transformer are strong and the noise levels in the MSE are low. The Transformer model is thus deployed in the application.

%

    % Author: Joel Janek Dabrowski
% Manuscript: Anomaly Detection for Prawn Farming
% Section: Conclusion

\section{Summary and Conclusion}
\label{sec:summaryAndConclusion}

We present a system which provides decision support through forecasting and anomaly detection. The forecasting models provide accurate forecasts with an average MAPE of $12\%$ over a 24-hour (96 step-ahead) forecast. We demonstrate that our novel approach to including weather forecast data into the models reduces both the forecast error and its variation. Two anomaly detection cases are presented demonstrating the ability to predict and identify a DO crash and sensor biofouling. The DO crash is detected 5 hours before the actual event.

Our system has been deployed in various phases since 2019 and we have received positive feedback from the farm. In future work, ensembles of models may provide more accurate forecasts and more refined anomaly detection. Furthermore, in order to capture more anomalies for further validation, more sensors will need to be deployed over a longer period of time.

\ifblindReviewRedaction

\else

\section{Acknowledgement}

Thanks to Pacific Reef Fisheries for providing us with the access to their farm to conduct this study and also for assisting us in deploying and maintaining sensors. This work was supported by the CSIRO Digiscape Future Science Platform.
\fi

    \bibliographystyle{splncs04}
    \bibliography{bibliography}
    
\end{document}